\documentclass[accepted]{uai2022} 

\usepackage[american]{babel}

\usepackage{natbib} 
\usepackage{mathtools} 
\usepackage{booktabs} 
\usepackage{tikz} 
\usepackage{amsfonts}
\usepackage{csquotes}
\usepackage{caption}
\usepackage{subcaption}\usepackage[ruled,linesnumbered]{algorithm2e}
\usepackage{amsmath}
\usepackage{amsthm}
\usepackage{amssymb}  
\usepackage{xcolor}
\usepackage{cancel}
\usepackage{times}
\usepackage{soul}
\usepackage{url}
\usepackage{bbm}
\DeclareMathOperator*{\argmax}{arg\,max}

\newcommand{\RSA}{\mathbb{R}^{|\mathcal{S}|\times|\mathcal{A}|}}

\newcommand{\RS}{\mathbb{R}^{|\mathcal{S}|}}

\newcommand{\AdaBracket}[1]{\left(#1\right)}
\newcommand{\AdaRectBracket}[1]{\left[#1\right]}

\newcommand{\AdaAngleProduct}[2]{\left\langle#1, #2\right\rangle}

\newcommand{\KL}{D_{\!K\!L}\!\AdaBracket{\pi || \bar{\pi}}}
\newcommand{\KLindex}[2]{D_{\!K\!L}\!\AdaBracket{\pi_{#1}|| \pi_{#2}}}
\newcommand{\KLany}[2]{D_{KL}\left(#1 \left| \! \right| #2 \right)}
\newcommand{\entropy}{\mathcal{H}\left( \pi \right)}
\newcommand{\entropyany}[1]{\mathcal{H}\left( #1 \right)}

\newcommand{\greedy}[1]{\mathcal{G}\!\AdaBracket{#1}}
\newcommand{\greedyH}[1]{\mathcal{G}^{\tau}\!\AdaBracket{#1}}
\newcommand{\greedyG}[1]{\mathcal{G}^{\alpha}\!\AdaBracket{#1}}
\newcommand{\approxgreedyG}[1]{\tilde{\mathcal{G}}^{\alpha}\!\AdaBracket{#1}}
\newcommand{\greedyN}{\mathcal{G}}
\newcommand{\greedyHN}{\mathcal{G}^{\tau}}
\newcommand{\greedyGN}{\mathcal{G}^{\alpha}}
\newcommand{\approxgreedyGN}{\tilde{\mathcal{G}}^{\alpha}}

\newcommand{\BellmanH}[1]{T^{\tau}_{\pi_{#1}}}
\newcommand{\BellmanG}[1]{T^{\alpha}_{\pi_{#1}}}
\newcommand{\Bellman}[1]{T_{\pi_{#1}}}
\newcommand{\approxBellmanG}[1]{\tilde{T}^{\alpha}_{\pi_{#1}}}
\newcommand{\eq}[1]{Eq.\,(#1)}

\title{Enforcing KL Regularization in  General Tsallis Entropy \\ Reinforcement Learning via Advantage Learning}

\author[1]{\href{mailto:<lingwei.andrew.zhu@gmail.com>?Subject=Your paper}{Lingwei~Zhu}{} }
\author[2]{Zheng~Chen}
\author[3]{Eiji~Uchibe}
\author[1]{Takamitsu~Matsubara}
\affil[1]{%
    Nara Institute of Science and Technology, Japan
}
\affil[2]{%
    Osaka University, Japan
}
\affil[3]{%
    Advanced Telecommunication Research, Japan
  }
  
\begin{document}
\maketitle

\begin{abstract}
  
  Maximum Tsallis entropy (MTE) framework in reinforcement learning has gained popularity recently by virtue of its flexible modeling choices including the widely used Shannon entropy and sparse entropy.
  However, non-Shannon entropies suffer from approximation error and subsequent underperformance either due to its sensitivity or the lack of closed-form policy expression.
  To improve the tradeoff between flexibility and empirical performance, we propose to strengthen their error-robustness by enforcing implicit Kullback-Leibler (KL) regularization in MTE motivated by Munchausen DQN (MDQN).
  We do so by drawing connection between MDQN and advantage learning, by which MDQN is shown to fail on generalizing to the MTE framework.
  The proposed method Tsallis Advantage Learning (TAL) is verified on extensive experiments to not only significantly improve upon Tsallis-DQN for various non-closed-form Tsallis entropies, but also exhibits comparable performance to state-of-the-art maximum Shannon entropy algorithms.

\end{abstract}
  
  \section{Introduction}
  \label{sec:introduction}

The successes of modern reinforcement learning (RL) rely crucially on implementing value-based methods with powerful function approximators \citep{mnih2015human,OpenAI2020}.
However, the susceptibility of value-based methods is also magnified: various sources of error can perturb action value estimates such that the non-optimal actions have temporally higher values than the optimal ones, putting unit probability mass on such deceptive actions can greatly slow down learning \citep{Fujimoto18-addressingApproximationError,fu2019-diagnosis} or even cause divergence \citep{Bertsekas:1996:NP:560669,wagner2011,Bellemare2016-increaseGap}.

\begin{figure}
    \centering
    \includegraphics[width=0.7\linewidth]{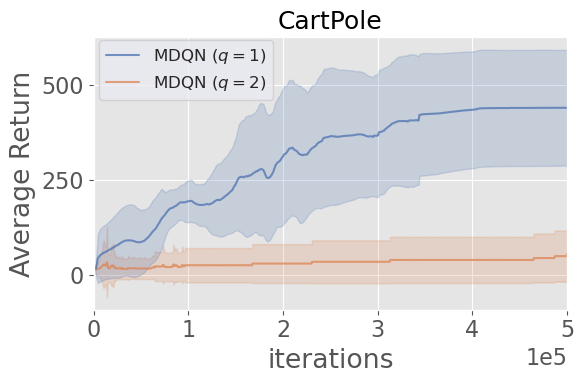}
    \caption{Munchausen DQN (MDQN) with entropic indices $q\!=\! 1,2$ (defined in Section \ref{sec:background}).
    For $q\!\neq\!1$ MDQN failed to learn anything on the simple environment \texttt{CartPole-v1}. }
    \label{fig:mt_cart}
\end{figure}

Recently there has seen promising progress towards alleviating this problem by leveraging \emph{entropy-regularized} algorithms \citep{haarnoja-SAC2018,ahmed19-entropy-policyOptimization}: by introducing entropy the policy becomes stochastic, hence optimal actions with temporarily low values can still be selected and subsequently help correct the value estimate. 
Besides the well-known Shannon entropy giving rise to Boltzmann softmax policy, the maximum Tsallis entropy (MTE) framework has gained recent popularity in that it offers more flexible modeling choices: by varying its entropic index $q$, Shannon entropy ($q\!=\!1$) and Tsallis sparse entropy ($q\!=\!2$) \citep{Martins16-sparsemax,Lee2018-TsallisRAL} can be recovered \citep{Lee2020-generalTsallisRSS}.
Other values between $1<q<2$ and $q>2$ can also serve as an interesting substitute to Shannon or sparse entropy for more compact/diverse policies. 
However, a common problem of those entropies is the lack of robustness against errors: they do not enjoy closed-form policy expression when $q\neq 1,2$, hence empirical underperformance is often incurred by significant greedy step error.
The sensitivity to errors holds true even for the closed-form sparsemax policy ($q\!=\!2$): due to its compactness, by contrast it is inherently less robust than the softmax \citep{chen2018-TsallisApproximate,Lee2020-generalTsallisRSS}.

Implicit Kullback-Leibler (KL) regularization might come to rescue for improving the error-robustness of non-Shannon entropies. 
Currently one of the state-of-the-art methods Munchausen Deep Q-Network (MDQN) \citep{vieillard2020munchausen} has demonstrated that significant performance boost can be attained by adding logarithm of the stochastic policy $\ln\pi$ as \emph{a bootstrapping term to any temporal difference learning algorithms for facilitating learning.}
The general claim and success of MDQN motivates us to investigate its Munchausen Tsallis DQN counterpart.
However, as shown in Figure \ref{fig:mt_cart}, MDQN typically fails to learn anything even for simple tasks with Tsallis entropy when $q\!\neq\!1$, which suggests that directly resorting to MDQN for Tsallis entropy is infeasible.

In this paper, we connect three relevant ideas: MDQN, KL regularization and advantage learning.
We show the implicit KL regularization of MDQN happens as a result of advantage learning which is due to the equivalence between log-policy and advantage function, which only holds for Shannon entropy $q\!=\!1$.
Therefore, the failure of MDQN in MTE is expected since  for $q \!\neq\! 1$ the equivalence is lost.
Motivated by this observation, we propose to enforce implicit KL regularization by explicitly enlarging the action gap in the MTE framework.
By extensive experiments we demonstrate that TAL achieves significant improvement over the state-of-the-art value-based MTE method Tsallis-DQN, even for those entropic indices that do not have closed-form policy expression. 
We also show its performance is competitive with state-of-the-art advantage learning algorithms (including original MDQN), which brings a competitive substitute for the conventional Shannon entropy in relevant domains \citep{Lee2020-generalTsallisRSS}.

The rest of the paper is organized as follows: in Section \ref{sec:related} we briefly survey related work.
In Section \ref{sec:background} we review the background for the proposed method in Section \ref{sec:approach}.
The proposed framework is comprehensively evaluated in Section \ref{sec:experiment}.
We conclude the paper in Section \ref{sec:discussion}.

\section{Related Work}\label{sec:related}

The very recent success of MDQN shows significantly improved performance can be attained by simply augmenting reward function with logarithm of the policy.
While the authors of \citep{vieillard2020munchausen} claimed such log-policy augmentation is generally applicable for any stochastic policy and demonstrated the effectiveness for soft Q-learning, we found that in practice MDQN worked poorly with the more general Tsallis entropy that is more suited to risk-sensitive applications \citep{Lee2018-TsallisRAL,Lee2020-generalTsallisRSS}, which begs the question of why MDQN achieved superior performance only with Shannon entropy.
The connection with action gap maximization was mentioned in \citep{vieillard2020munchausen}, though the authors attributed the superior performance of MDQN to implicit KL regularization.

The concept of advantage learning due to \citep{Baird1999-gradient-actionGap} lies in increasing the action gap and hence increasingly distinguishing the optimal action from suboptimal ones.
Besides rendering the algorithm empirically more robust, increased action gap comes with various benefits. 
For example, \cite{Farahmand2011-actionGap} showed that if the action gap regularity assumption holds, convergence rate of suboptimality gap towards zero can be much faster.
\cite{Bellemare2016-increaseGap} studied the non-stationary policy problem of Bellman optimality operator and proposed a solution by modifying based on the advantage learning scheme.
Recently, advantage learning was extended by \citep{Lu2019-generalStochasticAL} to the setting of random action gap coefficients and by \citep{Ferret2021-selfImitationAdvantage} to self-imitation learning.

The connection between advantage learning and entropy-regularized RL was first alluded to by \citep{azar2012dynamic} when studying KL-regularized policy iteration. 
This approach was followed by \citep{kozunoCVI} to incorporate Shannon entropy.
By defining a preference function, \cite{kozunoCVI} found that KL regularization is possible by iterating upon an advantage learning scheme.
In \citep{vieillard2020munchausen}, the authors showed how to derive CVI starting from MDQN. 
In this paper, we derive MDQN from CVI from an advantage learning perspective.
By connecting those ideas we justify the use of advantage learning for implicitly enforcing KL regularization in the MTE framework.

\section{Background}\label{sec:background}

\subsection{RL Basics}\label{sec:rl_basics}

Reinforcement learning problems are typically formulated by Markov decision processes (MDPs) which are expressed by the quintuple $(\mathcal{S}, \mathcal{A}, P, r,  \gamma)$, where $\mathcal{S}$ and $\mathcal{A}$ denote state space and action space, respectively. 
$|\mathcal{S}|$ defines the cardinality of state space and likewise for the action space.
$P$ denotes the transition probability kernel such that $P(\cdot|s,a)$ indicates the probability of transitioning to next state conditional upon taking action $a$ in state $s$. 
Let $r(s,a)$ denote the reward associated with that transition. When the time dependence $t$ should be made explicit, we write $r_t := r(s_t, a_t)$.
The discount factor $\gamma \!\in\! (0,1)$ trades off the importance between current and future rewards.
A policy $\pi$ maps a state to a distribution over actions.
We define the state-action value function $Q_{\pi}\in \RSA$ as:
\begin{align}
    Q_{\pi}(s,a) = \mathbb{E}\AdaRectBracket{\sum_{t=0}^{\infty} r_t | s_0=s , a_{0} = a} ,
    \label{eq:qvalue}
\end{align}
where the expectation is with respect to the trajectory induced by policy $\pi$ and transition probability $P$.

Value-based methods typically iterate on an initial $Q$ as:
$T_{\pi}Q \!:=\! r + \gamma P_{\pi}Q$, 
where $T_{\pi}$ denotes the Bellman operator and $\gamma P_{\pi}Q \!:=\! \gamma\mathbb{E}_{s'\sim P(\cdot|s,a),a'\sim\pi(\cdot|s')}\!\AdaRectBracket{Q(s',a')}$, with the definition being component-wise. 
Repeating the above iteration guarantees convergence to its unique fixed point $Q_{\pi}$.
Our goal is to find an optimal policy $\pi^*$ such that $T_{*}Q = \max_{\pi}T_{\pi}Q$.
The optimal policy can be extracted with the greedy operator $\pi(a|s)\!\!=\!\greedy{Q_{\pi}}\!:=\!\argmax_{\pi}Q_{\pi}$.
For convenience, for any two functions $F_1, F_2 \in\RSA$ we define their inner product as $\AdaAngleProduct{F_1}{F_2}\in\RS$.
We also define $\boldsymbol{1}$ as an all-one vector whose dimension should be clear from the context.

To render the policy robust against suboptimal actions that have temporally deceptively high values,  advantage learning scheme augments the standard Bellman operator $T_{\pi}Q$ with an additional action gap or advantage function: $A_{\pi}\in\RSA\!=\! Q_{\pi} - V_{\pi}$, where $V_{\pi} \!=\! \AdaAngleProduct{\pi}{Q_{\pi}}$ is the average of action values.
The agent is prompted to select actions that have values higher than the average to increasingly distinguish optimal actions from bad ones.

\subsection{Maximum Entropy RL}\label{sec:maxEnt}

In the recently popular maximum entropy RL literature, the reward is augmented by the Shannon entropy $\tau\entropy \!:=\! \tau\AdaAngleProduct{-\pi}{\ln\pi}$, where $\tau$ is the augmentation coefficient. 
By the Fenchel conjugacy \citep{geist19-regularized}, the Shannon entropy induced optimal policy is a Boltzmann distribution:
 \begin{align}
    \greedyH{Q} = \frac{\exp\AdaBracket{\tau^{-1}Q} }{\AdaAngleProduct{\boldsymbol{1}}{\exp\AdaBracket{\tau^{-1}Q}}} = \exp\AdaBracket{\tau^{-1}(Q-V)},
    \label{eq:boltzmann}
 \end{align} 
$V \!:=\! \tau\ln\AdaAngleProduct{\boldsymbol{1}}{\exp\AdaBracket{\tau^{-1}Q}}$ is the soft value function. 
 Its regularized Bellman operators are defined as $\BellmanH{} {Q}\!=\! r \!+\! \gamma P\!\AdaAngleProduct{\pi}{Q \!-\!\tau\ln\pi}$.

It turns out that the Shannon entropy is a special case of the well-known Tsallis entropy \citep{TsallisEntropy} parametrized by entropic index $q$:
\begin{align}
    \begin{split}
       \alpha H_q(\pi):= \alpha \frac{k}{q-1} \AdaBracket{1 - \AdaAngleProduct{\boldsymbol{1}}{\pi^q}},
    \end{split}
    \label{eq:tsallis_entropy}
\end{align}
where $\alpha$ is its regularization coefficient. 
The Shannon entropy is recovered by letting $k \!=\!\frac{1}{2}$ and $q\!\rightarrow\! 1$.
When $q\!\rightarrow \!\infty$, the regularization effect vanishes.
In this paper, we fix $k\!=\!\frac{1}{2}$ and consider various cases of $q$.
The corresponding regularized Bellman operator is hence given by $\BellmanG{}Q \!=\! r \!+\! \gamma P\AdaAngleProduct{\pi}{\!Q+\frac{\alpha}{2(q-1)}\!\AdaBracket{1-\pi^{q-1}}\!}$. 
Tsallis entropy induces the stochastic optimal policy as:
\begin{align}
    \greedyG{Q}(a|s) = \sqrt[\leftroot{-2}\uproot{16}q-1]{\AdaRectBracket{\frac{Q(s,a)}{\alpha} - \psi\AdaBracket{\frac{Q(s,\cdot)}{\alpha}}    }_{+} \frac{2(q-1)}{q}},
    \label{eq:sparse_policy}
\end{align}
where the operation $[\cdot]_{+}$ converts negative components to $0$.
$\psi$ is the normalization term to ensure the policy sums to one.
When $q\!=\!2$, we recover the sparsemax policy \citep{Lee2018-TsallisRAL,chow18-Tsallis} whose normalization is:
\begin{align}
    \begin{split}
        \psi\AdaBracket{\frac{Q(s,\cdot)}{\alpha}}  = \frac{\sum_{a\in S(s)} \frac{Q(s,a)}{\alpha} - 1 }{K{(s)}},
    \end{split}
    \label{eq:sparse_normalization}
\end{align}
where $S(s)$ is the set of actions satisfying $1 \!+\! i\frac{Q(s,a_{(i)})}{\alpha} \!>\! \sum_{j=1}^{i}\frac{Q(s,a_{(j)})}{\alpha}$, $a_{(j)}$ indicates the action with $j$th largest action value, $K(s)$ denotes the cardinality of $S(s)$.
The value function is 
\begin{align}
    V=\frac{1}{2}\sum_{j=1}^{K(s)}\AdaBracket{\AdaBracket{\frac{Q(s,a_{(j)})}{\alpha} }^2    -    \psi\AdaBracket{\frac{Q(s,\cdot)}{\alpha}}^2 } + \frac{1}{2}.
    \label{eq:sparse_value}
\end{align}
The sparsemax policy has been recently demonstrated as a suitable candidate for safety-sensitive tasks such as robotics \citep{Lee2018-TsallisRAL,Lee2020-generalTsallisRSS}.
However, the performance of its value-based methods typically cannot match Shannon entropy since it often results in insufficient exploration.

When $q\neq 1,2,\infty$, the normalization term and hence optimal policy might not have closed-form expression. 
But we can approximate the optimal policy by leveraging first order expansion following \citep{chen2018-TsallisApproximate}\footnotemark\footnotetext{\cite{chen2018-TsallisApproximate} considered the case of $k=1$, while we consider $k=\frac{1}{2}$ in this paper.}:
\begin{align} 
  \begin{split}
      \approxgreedyG{Q}(a|s) \approx \sqrt[\leftroot{-2}\uproot{16}q-1]{\AdaRectBracket{\frac{Q(s,a)}{\alpha} - \tilde{\psi}\AdaBracket{\frac{Q(s, \cdot)}{\alpha}}}_{+} },
  \end{split}
  \label{eq:approximate_policy}
\end{align}
where $\tilde{\psi}\AdaBracket{\frac{Q(s,\cdot)}{\alpha}}  \!\approx\! \frac{\sum_{a\in S(s)} \frac{Q(s,a)}{\alpha} - \frac{1}{2}q }{K{(s)}} + \AdaBracket{\frac{q}{2} - \frac{q}{q-2}}$ is the approximate normalization. The set $S(s)$ then allows actions satisfying $\frac{1}{2}q \!+\! i\frac{Q(s,a)}{\alpha} \!>\! \sum_{j=1}^{i}\!\!\frac{Q(s,a_{(j)})}{\alpha} \!+\! j(\frac{q}{2} \!-\! \frac{q}{q-2})$.
We denote its associated Bellman operator as $\approxBellmanG{}$.
Detailed derivation of approximate Tsallis policy is left to Appendix \ref{apdx:derivation}.

\section{Approach}\label{sec:approach}

\subsection{Munchausen RL}\label{sec:mdqn}

MDQN \citep{vieillard2020munchausen} proposes \emph{optimizing for the immediate reward augmented by the scaled log-policy of the agent when using any TD scheme}.
Specifically, MDQN was implemented based on soft Q-learning as follows:
\begin{align}
    \begin{split}
        \begin{cases}
            \pi_{k+1} = \argmax_{\pi} \AdaAngleProduct{\pi}{Q_{k}} + \tau\entropy & \\
            Q_{k+1} = r + {\color{red}{\beta\tau \ln\pi_{k+1} }} + \gamma P \AdaAngleProduct{\pi_{k+1}}{Q_k - {\color{blue}\tau\ln\pi_{k+1} } } &, 
        \end{cases}
    \end{split}
    \label{eq:mdqn_recursion}
\end{align}
where the red term is the \emph{Munchausen term}, the blue term comes from soft Q-learning.
For simplicity let $\beta=1$, MDQN has the following equivalence:
\begin{align}
    \begin{split}
    & Q_{k+1} - \tau\ln\pi_{k+1} = \\
    & \quad r + \gamma P\AdaBracket{\AdaAngleProduct{\pi_{k+1}}{Q_{k} - \tau\ln\pi_{k}} - \tau\KLindex{k+1}{k}} \\
    & \quad \Leftrightarrow Q'_{k+1} = r + \gamma P\AdaBracket{\AdaAngleProduct{\pi_{k+1}}{Q'_k} - \tau\KLindex{k+1}{k}},
\end{split}
\label{eq:mdqn}
\end{align}
where $Q'_{k} \!:=\! Q_{k} \!-\! \tau \ln\pi_{k}$ is the newly defined action value function.
Hence MDQN performs KL regularization by adding the Munchausen term $\ln\pi_{k+1}$ to soft Q-learning.
One of the crucial properties of KL regularization is its optimal policy averages over past action values: $\pi_{k+1}\propto\exp\AdaBracket{\tau^{-1}\sum_{j=1}^{k}Q_j}$ \citep{vieillard2020leverage}, which effectively cancels out errors under mild assumptions.

In the following section, we show the implicit KL regularization is performed when the Munchausen term $\ln\pi $ is equivalent to performing advantage learning under certain conditions, motivating our later proposal.

\subsection{An Advantage Learning Perspective for Munchausen RL}\label{sec:entropy_advantage}



It is worth noting that in the lieu of Shannon entropy augmentation, from \eq{\ref{eq:boltzmann}} one has $\ln\pi \!=\! \tau^{-1}\!\AdaBracket{Q-V}$, hence adding the Munchausen term coincides with advantage learning.
In the recent literature, conservative value iteration (CVI) \citep{kozunoCVI} comprehensively analyzes the relationship between advantage learning and KL regularization: CVI states the conditions under which advantage learning could implicitly perform KL regularization.

The connection between MDQN, KL regularization and advantage learning motivates us to investigate the possibility of enforcing implicit KL regularization for the MTE framework via advantage learning.
To do so, we need to verify that the superior performance of MDQN is due to advantage learning.
We re-derive MDQN by manipulating the CVI update rule since CVI generalizes advantage learning.
We provide full derivation in Appendix \ref{apdx:adv_kl} and succintly summarize the key steps here.

It turns out the CVI update rule can be written as the following, with $\Psi, W$ being the CVI action and state value function, respectively: 
\begin{align}
    \Psi_{k+1} = r + \gamma P \AdaAngleProduct{\pi_{k+1}}{\Psi_{k}} + \beta \AdaBracket{ \Psi_{k} - W_k},
\label{eq:cvi_valueiteration}
\end{align}
where $\beta \!\in\! [0,1]$ is a function of the Shannon entropy and KL divergence coefficients.

It is worth noting that \eq{\ref{eq:cvi_valueiteration}} is itself a new value iteration scheme \textbf{regardless of the definition} of $\Psi$ and $\pi$ since we can arbitrarily initialize $\Psi_{0}, \pi_{0}$.
CVI states that, if the policy $\pi_{k+1}$ in \eq{\ref{eq:cvi_valueiteration}} meets the Boltzmann assumption $\pi_{k+1}\propto\exp\AdaBracket{\tau^{-1}\Psi_k}$ (CVI itself uses Boltzmann policy but analyzes general advantage learning), then iterating upon \eq{\ref{eq:cvi_valueiteration}} along results in KL regularization.

The simplest method to ensure $\pi_{k+1}$ is Boltzmann is perhaps to perform soft Q-learning. 
By rewriting \eq{\ref{eq:cvi_valueiteration}} as a soft Q-learning scheme and replacing $\Psi, W$ with $Q,V$ we have:
\begin{align*}
    Q_{k+1} \!=\! r  +  {\color{red}{\beta\!\AdaBracket{ Q_{k} - V_k} }} + \gamma P \AdaAngleProduct{\pi_{k+1}}{Q_{k} \!-\! {\color{blue}{\tau\ln\pi_{k+1}} }},
\end{align*} 
where the blue part comes from soft Q-learning and the we recognize the red part is $\beta(Q_{k} - V_{k}) = \beta\tau\ln\pi_{k}$.
We hence see this scheme is same with the MDQN update rule \eq{\ref{eq:mdqn_recursion}}.

\begin{figure}
  \centering
  \includegraphics[width=0.9\linewidth]{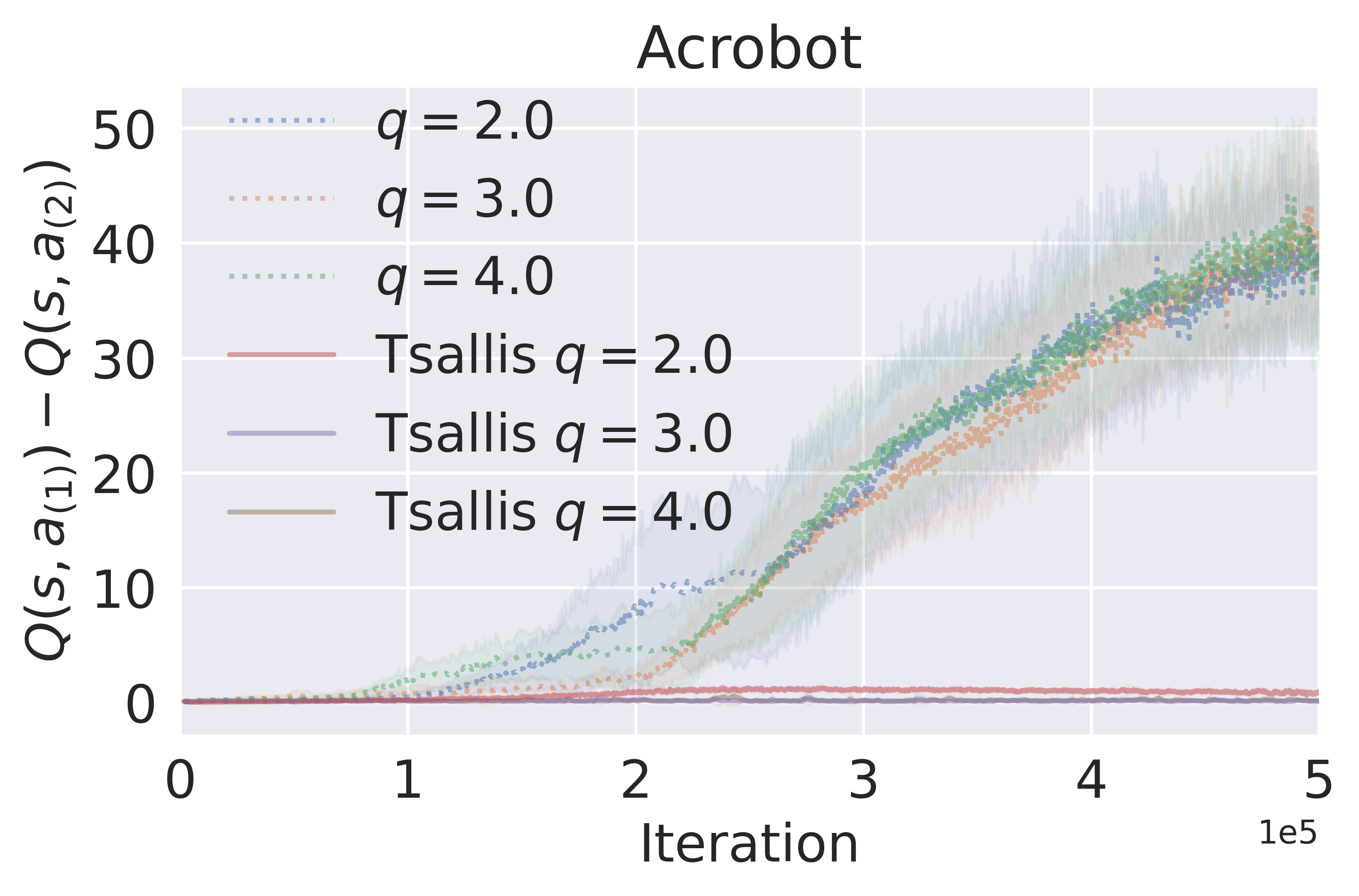}
  \caption{Action gap between the actions having the largest and second largest values $Q(s,a_{(1)}) - Q(s,a_{(2)})$ of TAL and Tsallis-DQN with different entropic indices.}
  \label{fig:action_gap}
\end{figure}

This above analysis implies that MDQN achieves implicit KL regularization and superior performance thanks to the equivalence between the Munchausen term and the action gap.
This intuition is confirmed by that in practice, MDQN actually computes the $\ln\pi$ term by $Q-V$ \citep[Appendix B.1]{vieillard2020munchausen}.
Hence as learning proceeds, the probabilities of optimal actions increase and so do $\ln\pi$, which implies the action gap is also increased as is shown in Figure \ref{fig:action_gap}.

The key for deriving the relationship between CVI and MDQN is the equivalence $\tau\ln\pi_{k+1} = Q_{k} - V_{k}$.
It is hence natural to conjecture that, if the equivalence does not hold as when $q\!\neq\!1$ in the general Tsallis entropy, MDQN can no longer perform implicit KL regularization, and learning performance might be significantly degraded.
As an example, consider \eq{\ref{eq:sparse_policy}} which is $q\!= \!2$.
The Munchausen term $\ln\pi \!=\! \ln\AdaRectBracket{\frac{Q(s,a)}{\alpha} - \psi\AdaBracket{\frac{Q(s,\cdot)}{\alpha}}}_{+}$ may be undefined since the policy assigns zero probabilities to some of the actions.
Even if we add some small value $\Delta$ in practice to prevent ill-defined $\ln\pi$, it does not encode action gap information since $\alpha\ln\pi =  \alpha\ln\AdaRectBracket{\frac{Q(s,a)}{\alpha} - \psi\AdaBracket{\frac{Q(s,\cdot)}{\alpha}} + \Delta }_{+}  \neq Q - V + \Delta$,
where $V$ was defined in \eq{\ref{eq:sparse_value}}.
This conclusion holds for other $q$ as well.
Hence it is confirmed that resorting to MDQN does not bring KL regularization for the MTE framework.

\begin{algorithm}[t]
    \caption{TAL-DQN}\label{algorithm}
    \SetKwInput{KwInput}{Input}
    \SetKwInput{KwInit}{Initialize}
    \KwInput{total steps $T$, update period $I$, \\ interaction period $C$, epsilon-greedy threshold $\epsilon$}
    \KwInit{ Network weights $\bar{\theta} = \theta$, \\ Replay buffer $B = \{\}$}
    \For{$t = 1, 2, \dots, T$ }
    {
        Collect tuple $(s_t, a_t, r_t, s_{t+1})$ in $B$ with $\pi_{q, \epsilon}$\\
        \If{ mod$(t, C) = 0$}
        {
            update $\theta$ with minibatch $B_t \subset B$ on $\mathcal{L}_{\texttt{TAL}}$ \\
        }   
        \If{mod$(t, I) = 0$}
        {
            update target network $\bar{\theta} \leftarrow \theta$\\
        }
    }
\end{algorithm}


\begin{figure*}[th!]
  \centering
  \includegraphics[width=\textwidth,trim=4 4 4 4,clip]{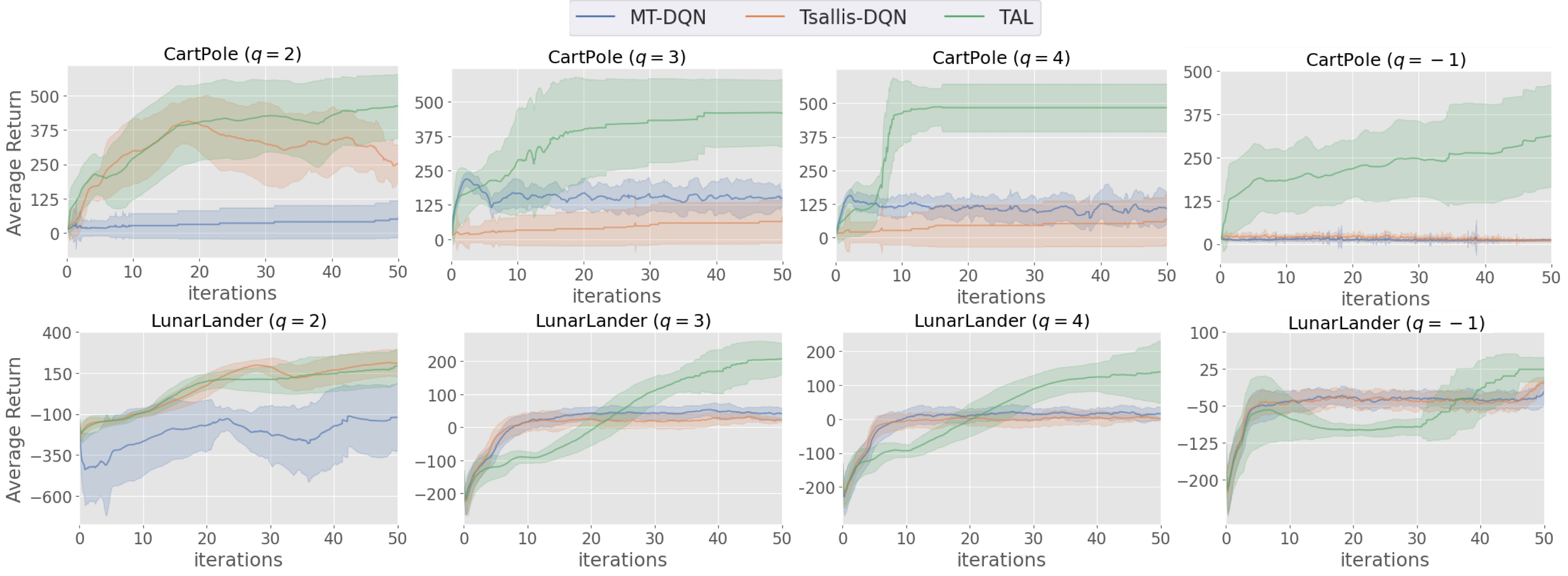}
  \caption{Comparison on \texttt{CartPole-v1} and \texttt{LunarLander-v2} between Tsallis Advantage Learning (TAL), Munchausen Tsallis DQN (MT-DQN) and Tsallis DQN with different entropic indices.
  All algorithms are averaged over 30 seeds to plot mean and $\pm 1$ standard deviation.
  For $q\!=\!-1,3,4$, approximate policy in \eq{\ref{eq:approximate_policy}} is employed. 
  When $q\!\leq\!0$ Tsallis entropy becomes convex, hence the regularizaton properties fail to hold.
  We show $q=-1$ for completeness.
  }
  \label{fig:cart_lunar_q2}
\end{figure*}

\begin{figure}[t]
  \centering
  \includegraphics[width=\linewidth,trim=4 4 4 4,clip]{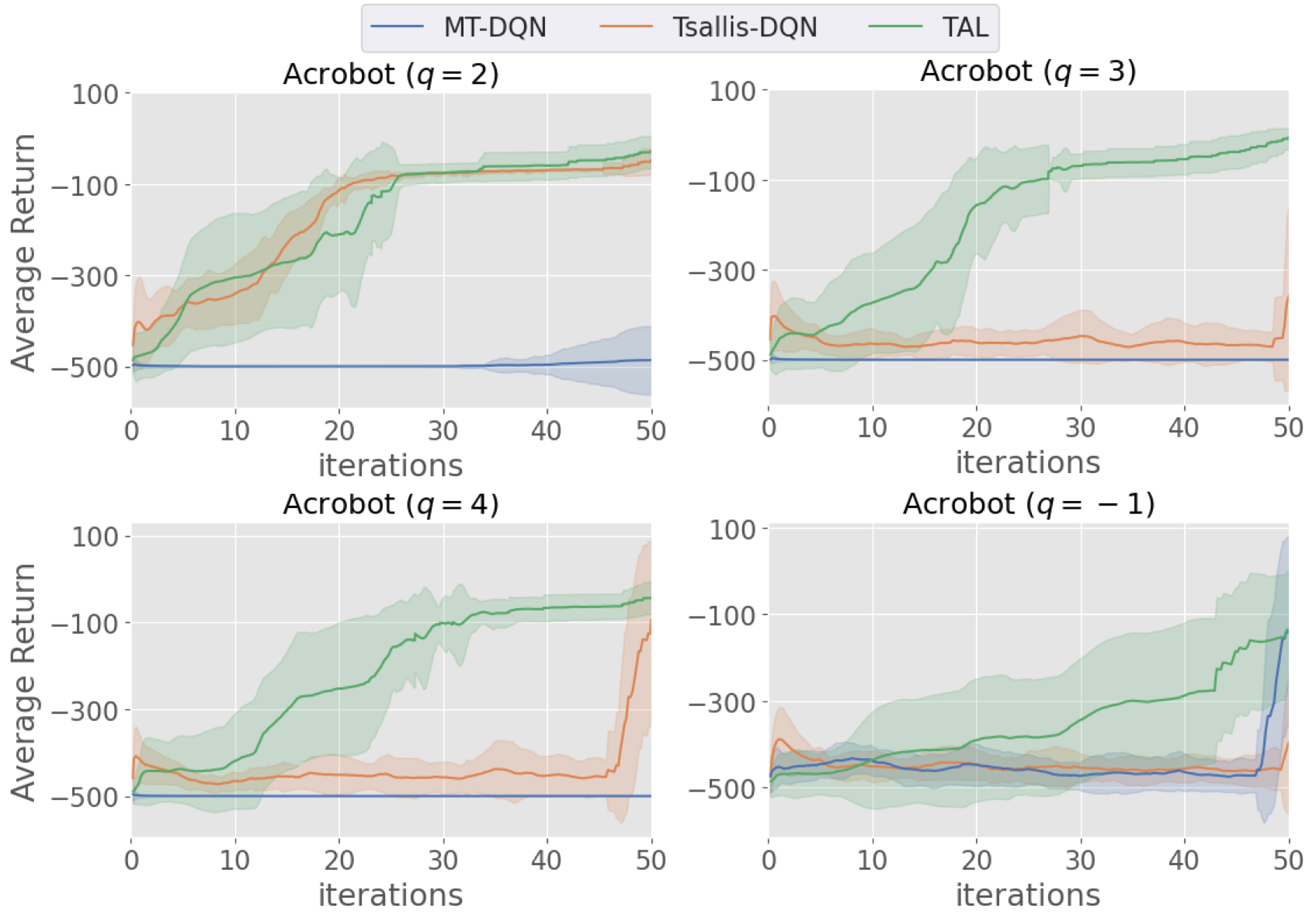}
  \caption{Comparison between Tsallis-DQN (orange), MT-DQN (blue) and TAL (green) with entropic indices $q\!=\! -1,2,3,4$ on \texttt{Acrobot-v1}. 
  }
  \label{fig:acrobot}
\end{figure}

\begin{figure}[t!]
  \centering
  \includegraphics[width=\linewidth,trim=4 4 4 4,clip]{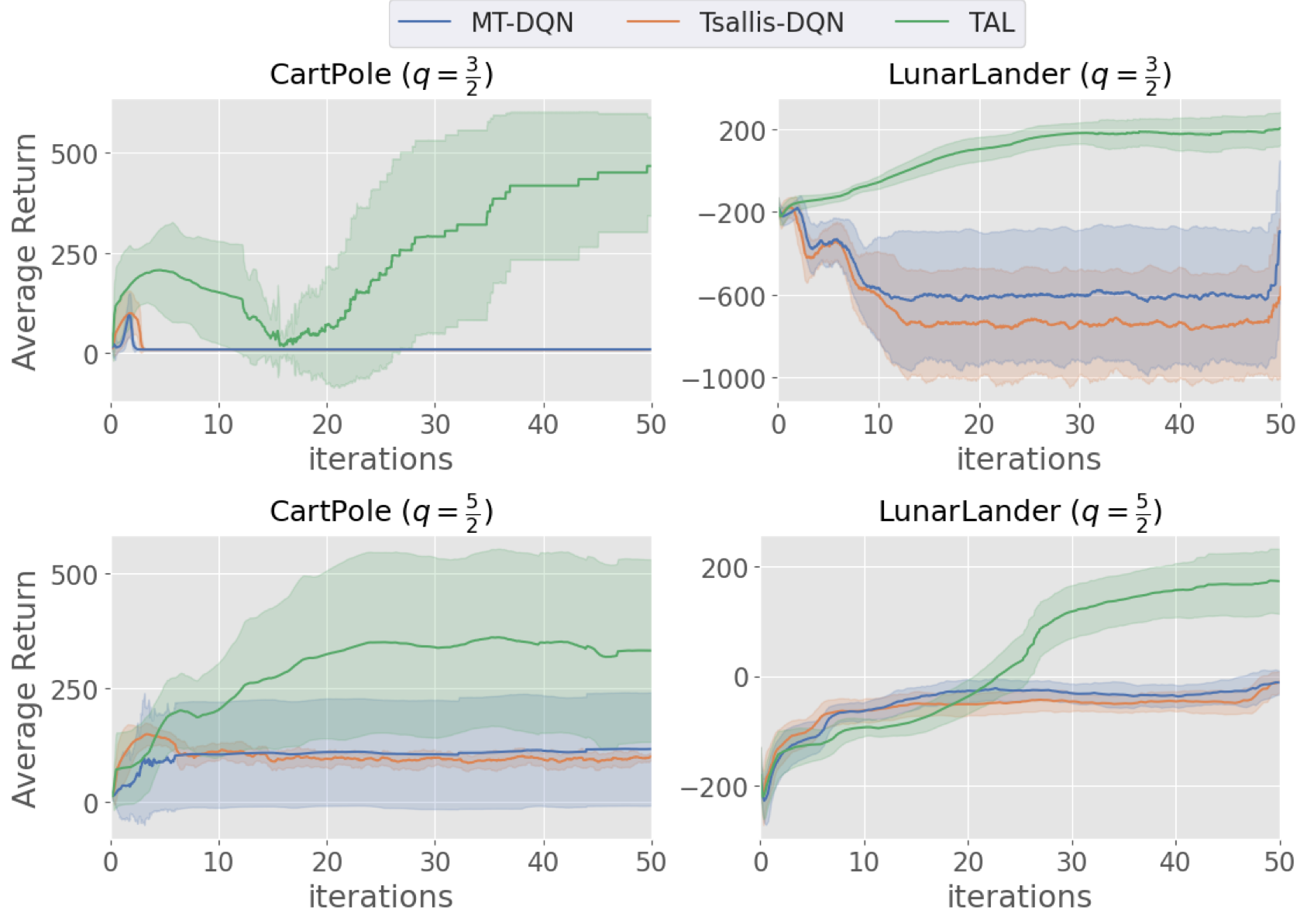}
  \caption{Comparison between Tsallis-DQN (orange), MT-DQN (blue) and TAL (green) with entropic indices $q\!=\! \frac{3}{2}, \,\, \frac{5}{2}$ on \texttt{CartPole-v1} and \texttt{LunarLander-v2}. 
  }
  \label{fig:cartLunar_fraction}
\end{figure}

\subsection{Advantage Learning for General Tsallis Entropy RL}

Motivated by the observation in the prior section, instead of relying on $\ln\pi$, we propose to explicitly perform advantage learning by increasing the action gap $Q - \AdaAngleProduct{\pi}{Q}$ to conform to the advantage learning scheme of \citep{azar2012dynamic,kozunoCVI}.
Though the policy with $q \!\neq\! 1$ is no longer Boltzmann, by \citep[Eq.\,(12)]{kozunoCVI} the Tsallis policy can still be used.
For justification the reader is referred to Appendix \ref{apdx:TAL}.

We can succinctly summarize the proposed method: Tsallis advantage learning  (TAL) as the following:
\begin{align}
    \begin{split}
        \begin{cases}
            \pi_{k+1} = \approxgreedyG{Q_k}, & \\
            Q_{k+1} = (\approxBellmanG{k+1})^{m} Q_{k} + \beta(Q_k - \AdaAngleProduct{\pi_{k+1}}{Q_k}), & \\ 
        \end{cases}
    \end{split}
    \label{eq:tsallis_advantage_learning}
\end{align}
where $\approxgreedyGN$ is the approximate Tsallis policy defined in \eq{\ref{eq:approximate_policy}}. 
We can also change the policy/Bellman operator pair $(\approxgreedyGN, \BellmanG{})$ to  $(\greedyHN, \BellmanH{}), (\greedyGN, \BellmanG{}), (\greedyN, \Bellman{})$ which correspond to settting $q\!=\!1, 2, \infty$, respectively.
When $q \!=\!1$, this scheme coincides with Munchausen RL.  When $q\!=\!\infty$, there is no regularization hence the scheme degenerates to advantage learning \citep{Baird1999-gradient-actionGap}.
The superscript $m$ indicates $m$ times application of the Bellman operator. Typically $m \!=\! 1, \infty$ which correspond to value iteration and policy iteration schemes. 
$\beta$ is the advantage coefficient and typically is between $[0,1]$.
$\AdaAngleProduct{\pi_{k+1}}{Q_k}$ is used to approximate the regularized value function $V$. We found that this approximation led to significantly better performance.

\begin{figure*}[t]
  \centering
  \includegraphics[width=\textwidth]{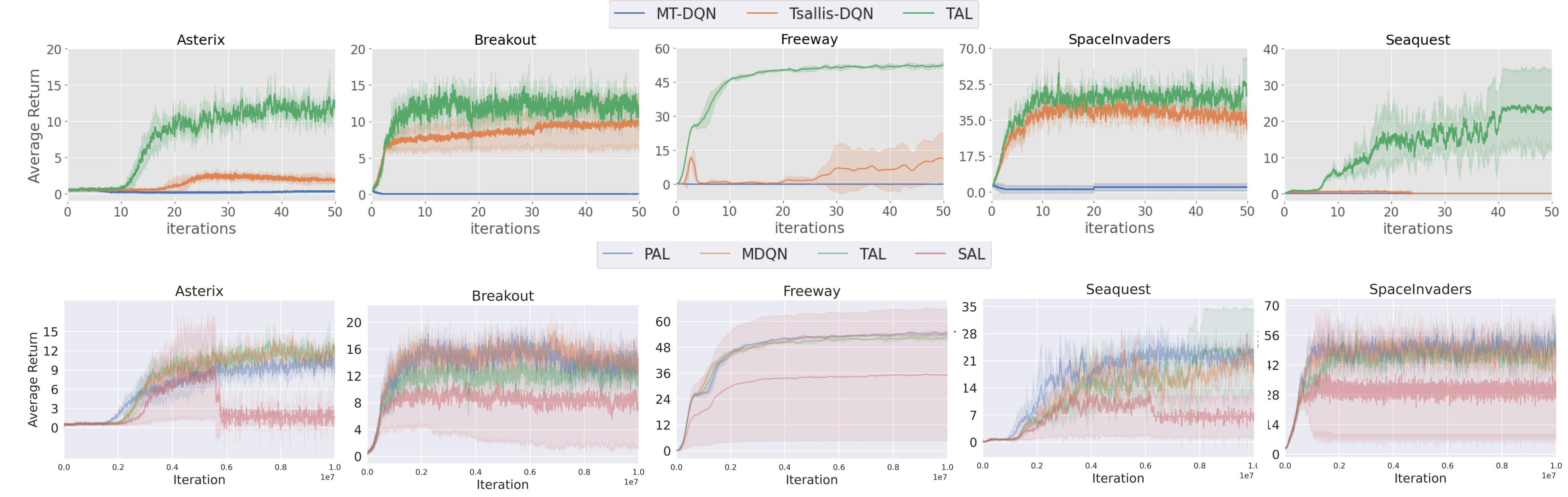}
  \caption{(Upper) Comparison between TAL, Tsallis-DQN and MT-DQN for $q\!=\!2$ on all MinAtar games. \\
  (Lower) Comparison between TAL, Munchausen-DQN (MDQN), Persistent AL (PAL) and Stochastic AL (SAL) on all MinAtar games. All algorithms are averaged over 3 seeds to plot mean and $\pm 1$ standard deviation.}
  \label{fig:minatar}
\end{figure*}

\begin{figure*}[th!]
  \centering
  \includegraphics[width=\textwidth,trim=4 4 4 4,clip]{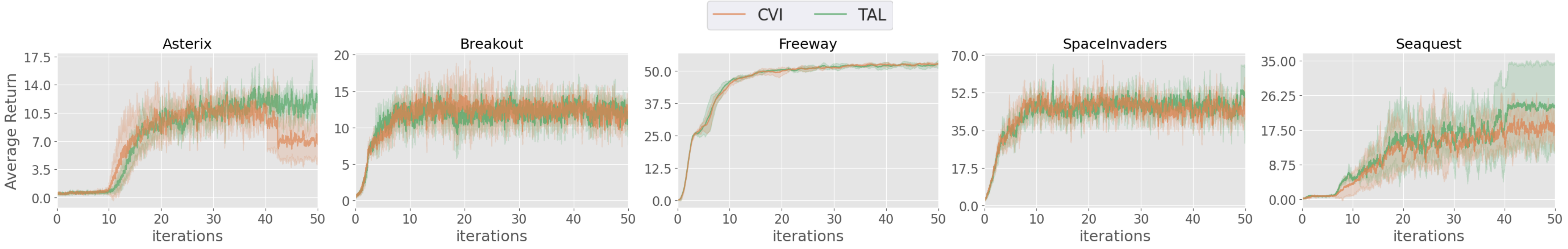}
  \caption{Comparison between conservative value iteration (CVI) and TAL ($q\!=\!2$) on all MinAtar games.
  The similar learning curves indicate TAL can achieve performance comparable to SOTA algorithms but with more desirable properties such as sparsity or parsimonious description of actions.
  }
  \label{fig:minatar_cvital}
\end{figure*}

In this paper we are interested in verifying effectiveness of the proposed method as a value iteration algorithm, i.e. $m=1$.
Specifically, we follow \citep{vieillard2020munchausen} to implement our algorithm based on the DQN architecture.
We maintain two networks with parameters $\theta, \bar{\theta}$, with the latter being target network updated after a fixed number of timesteps.
The loss function is defined by:
\[
    \mathcal{L}_{\texttt{TAL}}(\theta) \!:=\! 
    \hat{\mathbb{E}}_{B}\AdaRectBracket{ \AdaBracket{r \!+\! \gamma\AdaAngleProduct{\pi_{\bar{\theta}}}{Q_{\bar{\theta}}} \!+\! \beta\AdaBracket{Q_{\bar{\theta}} \!-\! \AdaAngleProduct{\pi_{\bar{\theta}}}{Q_{\bar{\theta}}}} \!-\! Q_{\theta} }^2 },
\]
where the policy $\pi_{\bar{\theta}}$ is computed following \eq{\ref{eq:tsallis_advantage_learning}}.
The algorithm Tsallis Advantage Learning - DQN (TAL-DQN) is listed in Alg. \ref{algorithm}.
We provide implementation details in Appendix \ref{apdx:implementation}.

\section{Experiments}\label{sec:experiment}

We aim to demonstrate that the proposed algorithm TAL is an effective generalization of Munchausen RL, in that when entropic index $q\!=\!1$ they coincide, but for other $q$ the Munchasuen bootstrapping term $\ln\pi$ can significantly deteriorate learning, while TAL can effectively boost the performance even for those $q$ that entail approximation.

For comprehensive evaluation, we select three different domains: simple Gym classic control \citep{brockman2016openai}, lightweight Atari games \citep{young19minatar} and challenging full-fledged Atari games \citep{bellemare13-arcade-jair}.
We aim to show TAL improves upon the SOTA value-based Tsallis entropy method Tsallis-DQN and achieves comparable performance with SOTA advantage learning methods. 

\subsection{Gym Classic Control}

We first examine the simple Gym classic control problems \texttt{CartPole-v1}, \texttt{LunarLander-v2} and \texttt{Acrobot-v1}. 
Their simplicity allows for evaluation with large number of seeds to verify our analysis. 
We name the algorithm Tsallis-DQN that sets $\beta=0$ in \eq{\ref{eq:tsallis_advantage_learning}} and compare Tsallis-DQN with TAL and Munchausen-Tsallis-DQN (MT-DQN) for entropic indices $q \!=\! -1, 2,3,4$.
When $q\!=\!2$ the policy has closed-form solution, for other indices  the approximation introduced in \eq{\ref{eq:approximate_policy}} is employed.
All algorithms are evaluated for $5\times 10^{5}$ steps with 30 seeds.
We decompose the $5\times 10^5$ steps into 50 iterations.
Implementation details such as network architecture and hyperparameters are provided in Appendix \ref{apdx:gym}.
The results are shown in Figure \ref{fig:cart_lunar_q2}.

Consistent with our analysis, Munchausen RL did not improve the performance for all entropic indices. 
For $q \!=\!2$, MT-DQN failed to learn any meaningful behavior even with analytic policy expression, while Tsallis-DQN performed relatively well. 
For $q\!=\! 3,4$, both MT-DQN and Tsallis-DQN failed to solve the \texttt{CartPole} environment.
It is also interesting to inspect the case of $q \!=\!-1$,  in which the Tsallis entropy becomes convex and hence the existing regularization results \citep{geist19-regularized,Li2019-regularizedSparse} fail to hold.
We show such result only for completeness, and one can expect the poor performance of all entropy-regularized algorithms. 
One can see the same trend holds true for \texttt{LunarLander-v2} and \texttt{Acrobot-v1} as well in Figure \ref{fig:acrobot}.
We quantitatively evaluate the action gap between the actions with the largest and second largest values, i.e. $Q(s,a_{(1)}) - Q(s,a_{(2)})$ and show the averaged results in Figure \ref{fig:action_gap}, which proves that the action gap is indeed enlarged.

One might also be interested in non-integer value of $q$, since we might benefit from the intermediate values from the two analytic choices $q\!=\!1,2$.
We therefore include the experiments with 
$q\!=\!  \frac{3}{2}, \frac{5}{2}$ and show them in Figure \ref{fig:cartLunar_fraction}.
It is apparent that that the trend still holds valid for non-integer values of $q$.
For $q \!=\!\frac{3}{2} \in [1,2]$,  we can expect this choice induces policy that stands between the softmax ($q\!=\!1$) and the Sparsemax ($q\!=\!2$), in that it is neither sparse nor assigns probability to all actions.
However, with this choice, on both environments Tsallis-DQN and MT-DQN performed extremely poor, leading to the lowest scores possible.
On the other hand, TAL managed to robustly converge to the optimal policy.

For $q\!=\!\frac{5}{2}$ the situation is similar, with small improvement of Tsallis-DQN and MT-DQN.
But we can see TAL did not converge to the optimal solution on \texttt{CartPole-v1}, this might be due to the hyperparameters in Table \ref{tb:gym} requires further fine-tuning for such specific choice.

\begin{figure*}[t]
  \centering
  \includegraphics[width=\textwidth,trim=4 4 4 4,clip]{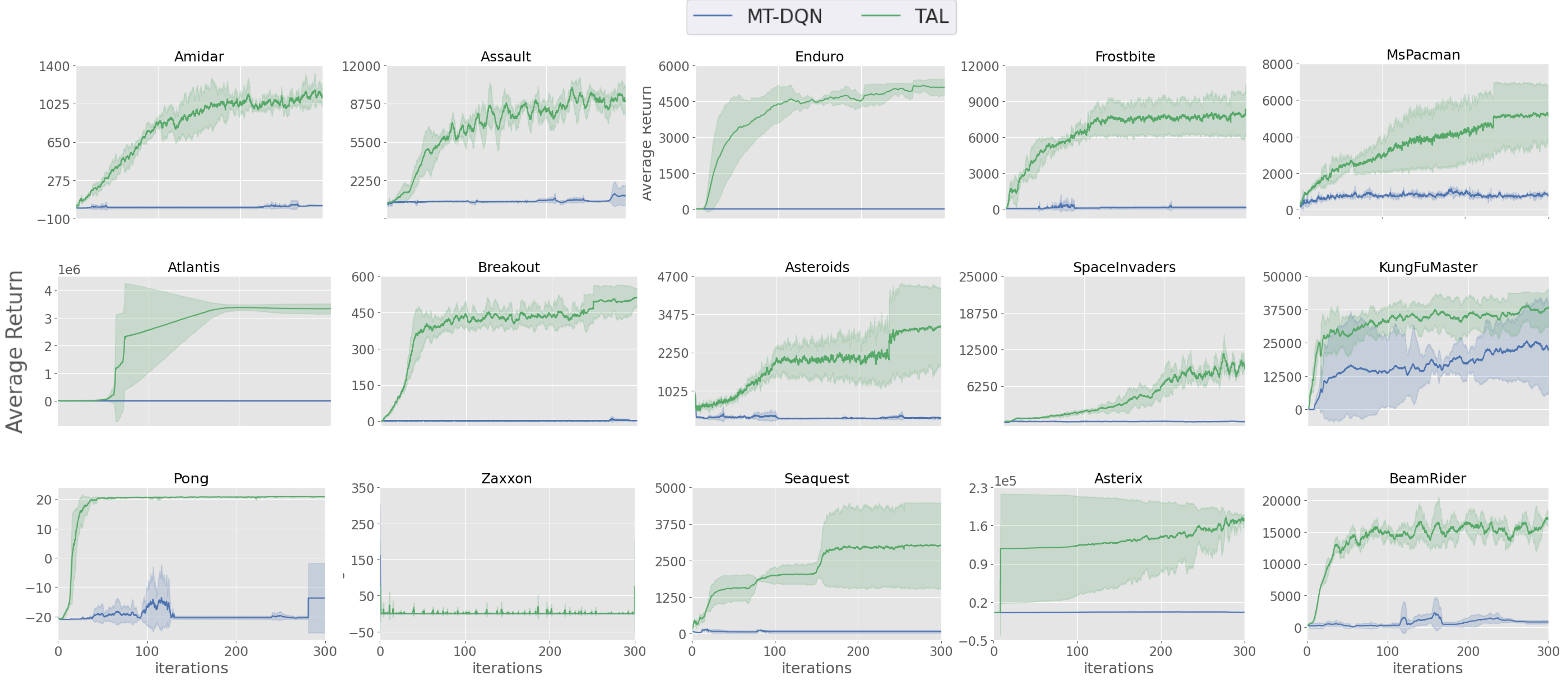}
  \caption{Comparison on full-fledged Atari games between TAL and MT-DQN  implemented based on QR-DQN for $q=2$. 
  }
  \label{fig:distributional_firstfive}
\end{figure*}

\begin{figure}[t]
  \centering
  \includegraphics[width=0.825\linewidth]{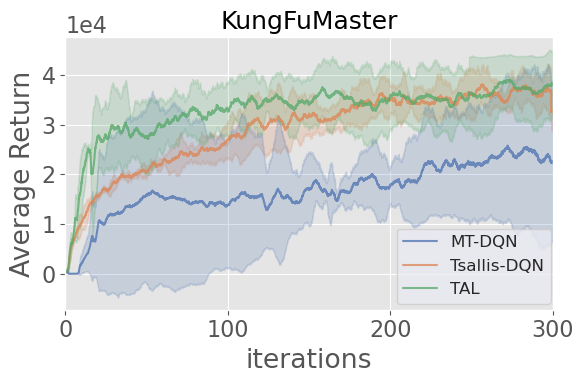}
  \caption{MT-DQN degraded the learning performance of Tsallis-DQN on the environment \texttt{KungFuMaster}. }
  \label{fig:kungfu}
\end{figure}

\begin{figure}[t]
  \centering
  \includegraphics[width=1.0\linewidth,trim=4 4 4 4,clip]{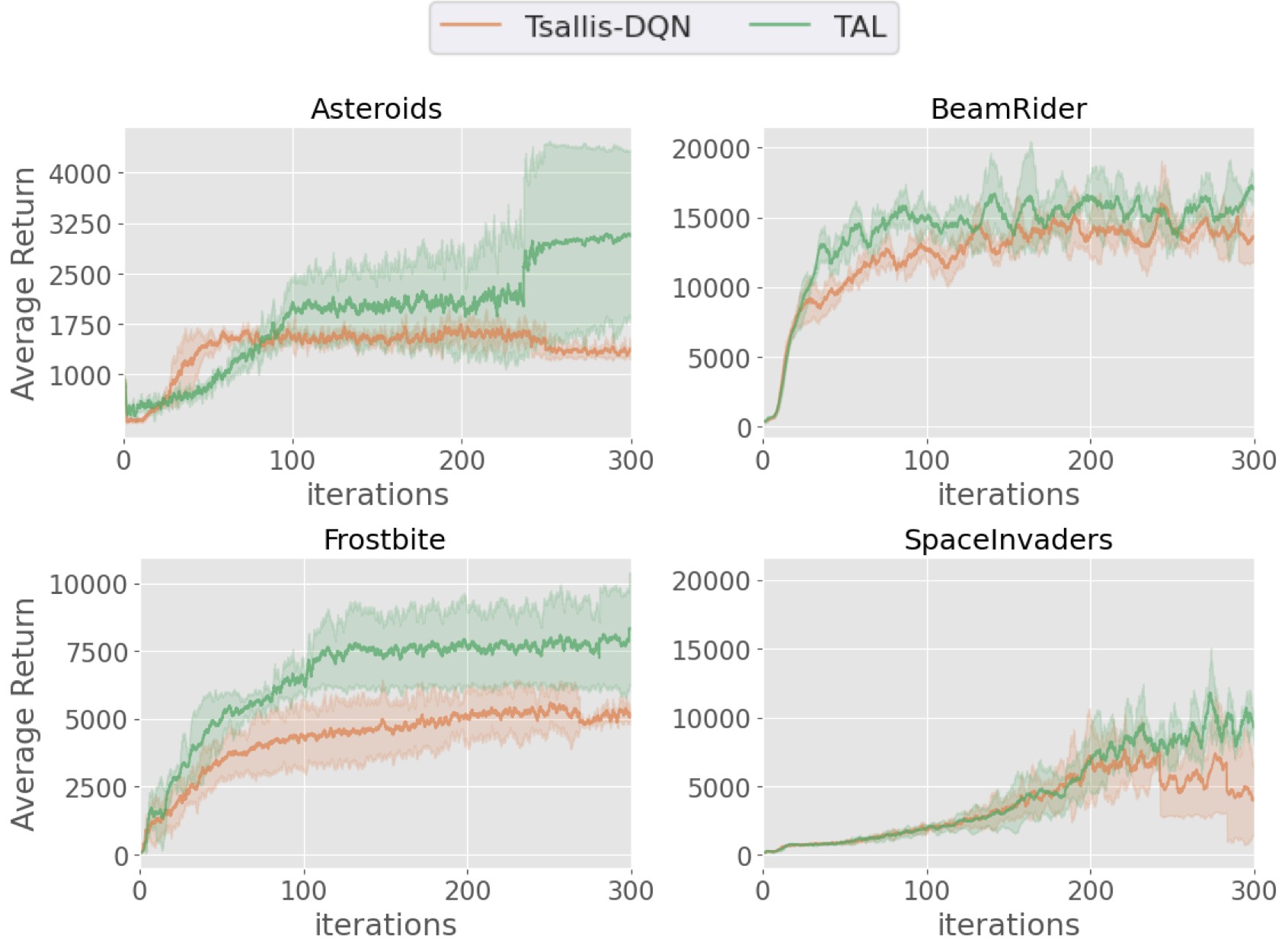}
  \caption{Comparison between TAL and Tsallis-DQN on four selected Atari games.}
  \label{fig:adv_improve}
\end{figure}

\subsection{MinAtar Games}

We inspect whether TAL can still boost the performance of Tsallis-DQN and work well on relatively complicated problems with high dimensional image as the state space. 
MinAtar provides 5 lightweight Atari games with optimization on multiple aspects including sticky actions, training frequency and reward normalization, etc.
We compare TAL with Tsallis-DQN, MT-DQN with entropic index $q\!=\!2$ since it enjoys closed-form policy expression.
 We also compare with persistent AL (PAL) \citep{Bellemare2016-increaseGap} that corrects the bias of the Bellman optimality operator; stochastic AL (SAL) that generalizes $\beta$ to random variables drawn from various distributions \citep{Lu2019-generalStochasticAL}. 
Every run consists of $10^{7}$ frames without frame skipping. 
We define every $200,000$ frames as one iteration so totally 50 iterations.
All algorithms are averaged over 3 seeds.
Since we use image as input, convolutional network is necessary. 
We detail the network architecture and hyperparameters used for all algorithms in Appendix \ref{apdx:minatar}.

We first examine whether TAL boosts the performance of Tsallis-DQN. 
From the upper row of Figure \ref{fig:minatar} it is clear that for relatively simple environments like \texttt{Breakout} and \texttt{SpaceInvaders}, TAL improved upon Tsallis-DQN slightly. 
However for challenging tasks like \texttt{Seaquest, Asterix, Freeway}, Tsallis-DQN failed to learn meaningful behavior while TAL successfully converged.
By contrast, MT-DQN showed flat learning curves for all environments, validating our analysis that $\ln\pi$ does not encode action gap information and hence can deteriorate learning.

We also ask whether TAL can compete with state-of-the-art advantage learning algorithms.
It is known that the hard max policy induced by Tsallis sparse entropy might cause brittle value function during learning, typically the performance of Tsallis entropy regularized algorithms cannot match SOTA algorithms such as PAL.
From the lower row of Figure \ref{fig:minatar} we can see that PAL and MDQN performed the best, but TAL had similar performance throughout.

On the other hand, SAL with the empirically performant choice of $\beta$ failed to keep up with other three algorithms due to the strong stochasticity in the advantage coefficient.
This section confirms that Tsallis entropy regularized algorithms can perform competitively with SOTA advantage learning algorithms while enjoying strong theoretical properties for policies, whose good performance was due to the advantage learning and subsequent implicit KL regularization.

\subsection{Distributional TAL on Atari Games}

Similar to Munchausen RL that can be combined with distributional RL methods. We demonstrate that TAL can also be leveraged in the context of distributional RL. 
Specifically, we implement TAL on top of the \emph{quantile regression DQN} (QR-DQN) \citep{Dabney2018-QRDQN}.

We compare TAL with MT-DQN with entropic index $q \!=\! 2$, since $q\!=\!1$ coincides with MDQN.
We choose a subset of 15 games from the full-fledged Atari games \citep{bellemare13-arcade-jair} for showing TAL remains effective for challenging problems.
Implementation details are provided in Appendix \ref{apdx:atari}.
Every single run consists of $3\times 10^7$ steps. We define every $10^{5}$ steps as one iteration so totally 300 iterations.
All algorithms are averaged over 3 seeds.

The learning results in Figure \ref{fig:distributional_firstfive} further confirmed our analysis that TAL is robust on complicated environments, and can work well with state-of-the-art RL algorithms.
By contrast, MT-DQN failed to learn any meaningful behaviors.

\textbf{Did TAL improve upon Tsallis-DQN? } 
We compare TAL and Tsallis-DQN on \texttt{KungFuMaster} in which MT-DQN showed non-trivial performance.
As illustrated by Figure \ref{fig:kungfu}, MT-DQN degraded the performance of Tsallis-DQN, proving the negative effect of the Tsallis Munchausen term.
In Figure \ref{fig:adv_improve} we compare Tsallis-DQN and TAL, and one can see that TAL still improved upon Tsallis-DQN, which is consistent with the trend shown in Gym and MinAtar environments.

\section{Discussion and Conclusion}\label{sec:discussion}

In this paper we proposed to implicitly perform KL regularization in the maximum Tsallis entropy framework to improve its error-robustness since the non-Shannon entropies are inherently less robust. 
We showed simply resorting to recently successful Munchausen DQN led to significantly degraded performance, due to the loss of equivalence between log-policy and advantage learning.
Accordingly, a novel algorithm Tsallis advantage learning was proposed to enforce implicit KL regularization.
By extensive experiments we verified that TAL improved upon the SOTA value-based MTE method Tsallis-DQN by a large margin, and exhibited comparable performance to SOTA advantage learning methods.

Several interesting future directions include combining TAL with actor-critic methods for further scalability,
and leveraging general logarithm function for the Munchausen term so the log-policy term again encodes advantage information.

\clearpage

\bibliography{../../library}

\clearpage

\appendix

\section{Derivation}\label{apdx:derivation}

\subsection{Derivation of Approximate Tsallis Policy}\label{apdx:approximate_policy}

The derivation of approximate Tsallis policy basically follows \citep{chen2018-TsallisApproximate} but with a general $k$ value instead of $k=1$.

The derivation begins with assuming an actor-critic framework where the policy network is parametrized by $w$.
It is well-known that the parameters should be updated towards the direction specified by the policy gradient theorem:
\begin{align}
    \Delta w \!\propto\! \mathbb{E}_{\pi, d^{\pi}} \!\! \AdaRectBracket{ Q_{\pi}\frac{\partial \ln\pi}{\partial w}  \!+\! \alpha \frac{\partial \entropy}{\partial w}  \!} \!-\! \sum_{s} \! \lambda(s) \frac{\partial\AdaAngleProduct{\boldsymbol{1}}{\pi}}{\partial w}  \!=:\! f(w),
    \label{eq:policy_network}
\end{align}
where $d^{\pi}$ denotes the stationary distribution induced by $\pi$, $\entropy$ denotes the Shannon entropy and $\alpha$ is the coefficient.
$\lambda(s)$ are the Lagrange multipliers for the constraint  $\AdaAngleProduct{\boldsymbol{1}}{\pi}=1$.
In the Tsallis entropy framework, we replace $\entropy$ with $H_q(\pi):=\frac{kq}{q-1}\AdaBracket{1 - \AdaAngleProduct{\boldsymbol{1}}{\pi^q}}$.

We can now explicitly write the optimal condition for the policy network parameters:
\begin{align}
    \begin{split}
        &f(w) \!=\! 0 \!=\!  \mathbb{E}_{\pi, d^{\pi}} \!\!\AdaRectBracket{ Q_{\pi}\frac{\partial \ln\pi}{\partial w} \!+\!  \alpha\frac{\partial H_q(\pi)}{\partial w} }  \!-\! \sum_{s} \lambda(s) \frac{\partial \AdaAngleProduct{\boldsymbol{1}}{\pi}}{\partial w} \\
        & = \mathbb{E}_{\pi, d^{\pi}}\!\! \AdaRectBracket{ Q_{\pi}\frac{\partial \ln\pi}{\partial w} - \alpha\frac{kq}{q-1}\AdaAngleProduct{\boldsymbol{1}}{\pi^{q}\frac{\partial \ln\pi}{\partial{w}}  } - \psi(s)\frac{\partial \ln\pi}{\partial w} } \\
        & = \mathbb{E}_{\pi, d^{\pi}}\!\! \AdaRectBracket{ \AdaBracket{ Q_{\pi} - \alpha\frac{kq}{q-1}{\pi^{q-1}  } - \psi(s) } \frac{\partial \ln\pi}{\partial w} },
    \end{split}
\end{align}
where we leveraged $\frac{\partial H_q(\pi)}{\partial w} = \frac{kq}{q-1}\AdaAngleProduct{\boldsymbol{1}}{\pi^{q}\frac{\partial \ln\pi}{\partial{w}}  }$ in the second step and absorbed terms into the expectation with respect to $\pi, d^{\pi}$ in the last step. 
$\psi(s)$ denotes the adjusted Lagrange multipliers by taking $\lambda(s)$ inside the expectation and modifying it according to the discounted stationary distribution.

Now it suffices to verify either $\frac{\partial \ln\pi}{\partial w}=0 $ or 
\begin{align}
    \begin{split}
        &Q_{\pi}(s,a) - \alpha\frac{kq}{q-1}{\pi^{q-1}(a|s)  } - \psi(s)  = 0\\
        \Leftrightarrow \quad &\pi^{*}(a|s) = \sqrt[\leftroot{-2}\uproot{16}q-1]{\AdaRectBracket{\frac{Q_{\pi}(s,a)}{\alpha}  - \psi\AdaBracket{\frac{Q_{\pi}(s,\cdot)}{\alpha}} }_{+}  \frac{q-1}{kq} }.
    \end{split}
\end{align}
When $q\!=\!2$, we recover the sparsemax policy \eq{\ref{eq:sparse_policy}} introduced in Section \ref{sec:maxEnt}.
When $q\!\neq\! 1,2$, the closed-form expression of $\pi, \psi$ might not exist.
Following \citep{chen2018-TsallisApproximate}, we leverage first order expansion on the policy as:
\begin{align}
    \begin{split}
        \tilde{\pi}^{*}(a|s) \approx 1 + \frac{1}{q-1}\AdaBracket{\AdaBracket{\frac{Q_{\pi}}{\alpha} - \tilde{\psi}\AdaBracket{\frac{Q_{\pi}}{\alpha}} } \frac{q-1}{kq}  - 1} .
    \end{split}
    \label{eq:apdx_approximate_policy}
\end{align}
By the constraint $\AdaAngleProduct{\boldsymbol{1}}{\pi} = 1$ we can approximately obtain the normalization as:
\begin{align}
    \begin{split}
        \tilde{\psi}\AdaBracket{\frac{Q(s,\cdot)}{\alpha}}  \approx \frac{\sum_{a\in S(s)} \frac{Q(s,a)}{\alpha} - kq }{K{(s)}} + \AdaBracket{kq - \frac{kq}{q-1}}.
    \end{split}
    \label{eq:apdx_approximate_normalization}
\end{align}
The associated set $S(s)$ of allowable actions must satisfy:
\begin{align}
    \begin{split}
       kq + i \frac{Q_{\pi}(s,a_{(i)})}{\alpha} \ge { \sum_{j=1}^{i}\frac{Q_{\pi}(s,a_{(j)})}{\alpha} } + j\AdaBracket{kq - \frac{kq}{q-1}}.
    \end{split}
\end{align}
In this paper, we consider the case $k\!=\!\frac{1}{2}$, which leads to the following approximate policy and normalization:
\begin{align} 
    \begin{split}
        \tilde{\psi}\AdaBracket{\frac{Q(s,\cdot)}{\alpha}}  \approx \frac{\sum_{a\in S(s)} \frac{Q(s,a)}{\alpha} - \frac{1}{2}q }{K{(s)}} + \AdaBracket{\frac{q}{2} - \frac{q}{q-2}},
    \end{split}
\end{align}
\begin{align}
    \begin{split}
        \approxgreedyG{Q} &\approx 1 \!+\! \frac{1}{q-1}\AdaBracket{\!\AdaBracket{\frac{Q}{\alpha} - \tilde{\psi}\AdaBracket{\frac{Q}{\alpha}\!}\!}\frac{2(q-1)}{q} - 1}     \\
         &\approx \sqrt[\leftroot{-2}\uproot{16}q-1]{\AdaRectBracket{\frac{Q}{\alpha} - \tilde{\psi}\AdaBracket{\frac{Q}{\alpha}}}_{+} }.
    \end{split}
\end{align}
where the set $S(s)$ then allows actions satisfying $\frac{1}{2}q \!+\! i\frac{Q(s,a)}{\alpha} \!>\! \sum_{j=1}^{i}\!\!\frac{Q(s,a_{(j)})}{\alpha} \!+\! j(\frac{q}{2} \!-\! \frac{q}{q-2})$.
We denote its associated Bellman operator as $\approxBellmanG{}$.

\subsection{Advantage Learning as KL Regularization}\label{apdx:adv_kl}
This section shows our derivation of connection between CVI and MDQN which is different from the connection shown in \citep[Appendix A.4]{vieillard2020munchausen}, but our conclusions are essentially the same.

Conservative Value Iteration (CVI) \citep{kozunoCVI} investigates the scheme maximizing:
\begin{align}
    \begin{split}
        \mathbb{E}\AdaRectBracket{\sum_{t=0}^{\infty} \gamma^t \AdaBracket{ r_t + \tau\entropyany{\pi(\cdot|s_t)} - \sigma \KLany{\pi(\cdot|s_t)}{\bar{\pi}(\cdot|s_t)}} },
    \end{split}
\end{align}
where $\entropy \!:=\! -\AdaAngleProduct{\pi}{\ln\pi}$ is the Shannon entropy and $\KL \!:=\! \AdaAngleProduct{\pi}{\ln\pi-\ln\bar{\pi}}$ is the KL divergence.
By the Fenchel conjugacy, we can show that the optimal policy takes the form
\begin{align}
    \pi^* \propto \bar{\pi}^{\frac{\sigma}{\sigma+\tau}}\exp\AdaBracket{\frac{1}{\sigma+\tau} \AdaBracket{r + \gamma P V^{*}_{\bar{\pi}}}},
    \label{eq:cvi_policy}
\end{align}
where $ V^{*}_{\bar{\pi}}$ is the regularized value function.
In practice, at the $k\!+\!1$-th iteration, $\bar{\pi}$ is set to the previous policy $\pi_{k}$.
Hence \eq{\ref{eq:cvi_policy}} can be written as $\pi_{k+1}\propto {\pi_k}^{\frac{\sigma}{\sigma+\tau}}\exp\AdaBracket{\frac{1}{\sigma+\tau}  \AdaBracket{r + \gamma P V_{{\pi_k}} }}$.
While a two-loop policy iteration algorithm can be employed to obtain $\pi_{k+1}$, we can avoid computing the  recursively defined $\pi_{k}$ by defining the action preference function:
\begin{align}
    \Psi_{k+1} := r + \gamma PV_{\pi_{k}} + \sigma \ln{\pi_{k}} = Q_{k+1} + \sigma \ln{\pi_{k}}.
\end{align}
Now the optimal policy for $k\!+\!1$-th iteration can be compactly expressed as 
\begin{align}
    \pi_{k+1} = \frac{\exp\AdaBracket{ \frac{1}{\sigma+\tau} \Psi_{k+1} }}{\AdaAngleProduct{\boldsymbol{1}}{\exp\AdaBracket{\frac{1}{\sigma+\tau}\Psi_{k+1}}}}.
\end{align}
CVI converges to the maximum Shannon entropy optimal policy by iterating upon $\Psi$:
\begin{align}
    \Psi_{k+1} = r + \gamma P \AdaAngleProduct{\pi_{k}}{\Psi_{k}} + \frac{\sigma}{\sigma+\tau} \AdaBracket{ \Psi_{k} - W_{k}},
    \label{eq:cvi_recursion}
\end{align}
where $W_k$ is the value function associated with $\Psi_{k}$ and $\pi$ is a Boltzmann policy.
\eq{\ref{eq:cvi_recursion}} implies that CVI performs a recursion which evaluates the preference function first: this suggests CVI is different from the conventional value iteration methods such as \eq{\ref{eq:mdqn_recursion}}:
\begin{equation}
    \begin{split}
        \begin{cases}
            \Psi_{k+1} = r + \gamma P \AdaAngleProduct{\pi_{k}}{\Psi_{k}} + \frac{\sigma}{\sigma+\tau} \AdaBracket{ \Psi_{k} - W_k} &, \\
            \pi_{k+1} = \exp\AdaBracket{\frac{1}{\sigma+\tau}\Psi_{k+1}} / \AdaAngleProduct{\boldsymbol{1}}{\exp\AdaBracket{\frac{1}{\sigma+\tau}\Psi_{k+1}}} &.
        \end{cases}
    \end{split}
    \label{eq:cvi_iteration}
\end{equation}
Without loss of generality we can reverse the order in \eq{\ref{eq:cvi_iteration}} by assuming $\pi_0$ is a uniform policy:
\begin{align}
    \begin{split}
        \begin{cases}
            \pi_{k+1} = \exp\AdaBracket{\frac{1}{\sigma+\tau}\Psi_{k}} / \AdaAngleProduct{\boldsymbol{1}}{\exp\AdaBracket{\frac{1}{\sigma+\tau}\Psi_{k}}}, & \\
            \Psi_{k+1} \!=\! r \!+\! \gamma P \AdaAngleProduct{\pi_{k+1}}{\Psi_{k}} \!+\! \frac{\sigma}{\sigma+\tau} \AdaBracket{ \Psi_{k} \!-\! W_k}, &
        \end{cases}
    \end{split}
    \label{eq:reverse_cvi}
\end{align}
where the definition of the action preference function is modified as:
\begin{align}
    \Psi_{k+1} :=  Q_{k+1} + \sigma \ln{\pi_{k+1}}.
    \label{eq:new_pref}
\end{align}
We see if we set $\beta :=\frac{\sigma}{\sigma+\tau}\in[0,1]$ and $\zeta := \frac{1}{\sigma + \tau} \in (0,\infty]$ in \eq{\ref{eq:reverse_cvi}}, we obtain \eq{\ref{eq:cvi_valueiteration}}.
Hence the CVI recursion can now be rewritten as:
\begin{align}
    \Psi_{k+1} = r + \gamma P \AdaAngleProduct{\pi_{k+1}}{\Psi_{k}} + \beta\AdaBracket{ \Psi_{k} - W_{k}}.
    \label{eq:new_cvi_recursion}
\end{align}    
It is worth noting that \eq{\ref{eq:new_cvi_recursion}} is itself a new value iteration scheme, regardless of the definition of $\Psi$.
CVI states that, if the policy satisfies the Boltzmann assumption, then simply iterating upon \eq{\ref{eq:new_cvi_recursion}} results in KL regularization.

The above analysis implies that any value iteration methods could enjoy the KL regularization properties by replacing $\Psi, W$ with $Q,V$  without needing to care about the definition of $\Psi, W$ since we can arbitrarily initialize $Q_0\!=\!\Psi_0\!=\!0$.
The simplest method to ensure the policy meets the Boltzmann assumption is by performing soft Q-learning on top of \eq{\ref{eq:new_cvi_recursion}}.
By rewriting \eq{\ref{eq:new_cvi_recursion}} as a soft Q-learning scheme and replacing every appearance of $\Psi, W$ with $Q, V$, we have:
\begin{align}
    Q_{k+1} \!=\! r  +  {\color{red}{\beta\!\AdaBracket{ Q_{k} - V_k} }} + \gamma P \AdaAngleProduct{\pi_{k+1}}{Q_{k} \!-\! {\color{blue}{\tau\ln\pi_{k+1}} }},
\end{align}
where the blue part is due to the definition of soft Q-learning as in \eq{\ref{eq:mdqn_recursion}} to ensure the policy is a Boltzmann distribution.
By observing that $\beta\!\AdaBracket{ Q_{k} - V_k} = \beta\tau\ln\pi_{k}$ we see this is the Munchausen DQN update rule \eq{\ref{eq:mdqn_recursion}}.

The key for deriving the relationship between CVI and MDQN is the equivalence $\tau\ln\pi_{k+1} = Q_{k} - V_{k}$.
It is hence natural to conjecture that, if the equivalence does not hold as when $q\!\neq\!1$ in the general Tsallis entropy, MDQN can no longer perform implicit KL regularization.
Furthermore, the contraction property might be lost by adding such $\ln\pi$ term.

\subsection{Tsallis Advantage Learning for KL Regularization}\label{apdx:TAL}

In Section \ref{apdx:adv_kl} we analyzed the relationship between advantage learning and KL regularization. 
In practice, the value function $W_k$ in CVI recursion \eq{\ref{eq:cvi_recursion}} is approximated by $\AdaAngleProduct{\pi_k}{\Psi_k}$ as it saves computation and achieves numerical stability \citep{azar2012dynamic,kozunoCVI}.
The requirement of $\pi$ being Boltzmann distribution is not a must, as can be seen from \citep[Eq.\,(12)]{kozunoCVI} which states that any policy $\mu_k$ satisfying the relation $\AdaAngleProduct{\mu_{k}}{\Psi_{k}} \geq \AdaAngleProduct{\pi_{k}}{\Psi_{k}}$ can be used.
This constitutes the theoretical basis for our use of $\AdaAngleProduct{\pi_k}{Q_k}$ in \eq{\ref{eq:tsallis_advantage_learning}} at the absence of analytic value function expression. 
Even for $q\!=\!2$ where the closed-form $V$ is available, we found that using the advantage as $Q_k - \AdaAngleProduct{\pi_k}{Q_k}$ significantly improved the performance and stability.


\section{Implementation Details}\label{apdx:implementation}

In this section we provide implementation details for all environments we used. 
Specifically, three different architectures were leveraged for handling gym environments, MinAtar games and Atari games, detailed in Sections \ref{apdx:gym}, \ref{apdx:minatar} and \ref{apdx:atari}, respectively.
We added a small number $\Delta$ to $\pi$ to ensure $\ln\pi$ is well defined.
For fast and reproducible evaluation, we implemented all algorithms based on the Stable-Baselines3 library \citep{stable-baselines3}.

\subsection{Gym Classic Control }\label{apdx:gym}

We introduce the parameters we used for the Gym environments in this section. 
Specifically, the hyperparameters are listed in Table \ref{tb:gym}.
The epsilon threshold is fixed at $0.01$ from the beginning of learning. 
The Munchausen number $\Delta$ refers to the small value added to the policy to prevent ill-defined $\ln\pi$.
FC$\,n$ refers to the fully connected layer with $n$ activation units.

\begin{table}[h!]
    \centering
    \caption{Parameters used for TAL, MT-DQN and Tsallis-DQN on \texttt{CartPole-v1} and \texttt{LunarLander-v2}. }
    \label{tb:gym}
    \begin{tabular}{llr}
        \cline{1-2}
        Parameter    & Value \\
        \hline
        $T$ (total steps)     &  $5\times 10^5$     \\
        $C$ (interaction period) &      4      \\
        $|B|$ (buffer size)      & $5 \times 10^4$     \\
        $B_{t}$ (batch size)  &  128 \\
        $\gamma$ (discount rate)       & 0.99      \\
        $I$  (update period)      & $1000$ (Car.) / $2500$ (Others) \\
        $\epsilon$ (epsilon greedy threshold) & 0.01 \\
        $\Delta$ (Munchausen number) & $10^{-8}$\\
        $\alpha$ (Tsallis entropy coefficient) &  0.03 \\
        $\beta$ (advantage coefficient) & 0.99 \\
        Q-network architecture & FC512 - FC512 \\
        activation units & ReLU \\
        optimizer & Adam \\
        optimizer learning rate & $10^{-3}$ \\
        \hline
    \end{tabular}
\end{table}

\subsection{MinAtar Games}\label{apdx:minatar}

For MinAtar games we employed the configuration recommended by \citep{young19minatar}.
The hyperparameters are listed in Table \ref{tb:minatar}.

Since MinAtar games optimize different aspects of Atari games, no frame skipping is necessary and hence the for every frame we update the network. 
The epsilon greedy threshold $\epsilon: 1.0 \rightarrow 0.05 |_{10\%}$ denotes that $\epsilon$ is initialized as 1.0 and gradually decays to $0.05$ through the first 10\% of learning.

The parameter $\tau$ is the Shannon entropy coefficient used for MDQN and plays a similar role as $\alpha$. 
The advantage learning coefficient $\beta$ is also used for PAL.
For SAL, we use the empirically most performant choice $\beta_{\texttt{SAL}}$ of \citep{Lu2019-generalStochasticAL} which is drawn from the Beta$(2,9)$ distribution and adds a constant $\frac{7}{9}$.
Then $\mathbb{E}\AdaRectBracket{\beta_{\texttt{SAL}}} = 1$ and $\mathbb{V}\mathbf{ar}\AdaRectBracket{\beta_{\texttt{SAL}}} = \frac{7}{405}$.

The network architecture consists of only one convolutional layer where $\text{Conv}^{d}_{a,b}c$ denotes a convolutional layer with $c$ filters of size $a\times b$ and stride $d$.

In Appendix \ref{apdx:adv_kl}  we analyzed the relationship between MDQN and CVI and in Figure \ref{fig:minatar_cvital}
we compared TAL against CVI and TAL with entropic index $q\!=\!2$ to see if TAL can achieve comparable performance to the state-of-the-art algorithm. 
CVI used the same Boltzmann temperature $\frac{1}{\sigma+\tau} =0.03$  and advantage coefficient $\beta$ as in Table \ref{tb:minatar}.
Since CVI and TAL exhibited similar performance, it might be safe to replace the Shannon entropy with Tsallis entropy in many applications where safety or parsimonious description of the action space are necessary \citep{Lee2018-TsallisRAL}, since the maximum Shannon entropy policy always assigns non-negligible probabilities to sub-optimal or even unsafe actions.


\begin{table}
    \centering
    \caption{Parameters used for TAL, MT-DQN, Tsallis-DQN, PAL, SAL and MDQN on all MinAtar games. }
    \label{tb:minatar}
    \begin{tabular}{llr}
        \cline{1-2}
        Parameter    & Value \\
        \hline
        $T$ (total steps)     &  $1 \times 10^7$     \\
        $C$ (interaction period) &      1      \\
        $|B|$ (buffer size)      & $1 \times 10^5$     \\
        $B_{t}$ (batch size)  &  32 \\
        $\gamma$ (discount rate)       & 0.99      \\
        $I$  (update period)      & $1000$  \\
        $\epsilon$ (epsilon greedy threshold) & $1.0 \rightarrow 0.05 |_{10\%}$ \\
        $\Delta$ (Munchausen number) & $10^{-8}$\\
        $\alpha$ (Tsallis entropy coefficient) &  0.03 \\
        $\tau$ (Munchausen entropy coefficient)  & 0.03 \\
        $\beta$ (advantage coefficient) & 0.9 \\
        $\beta_{\texttt{SAL}}$ (SAL advantage coefficient) & Beta(2, 9) + $\frac{7}{9}$\\
        Q-network architecture & $\text{Conv}^{1}_{3,3}16$ - FC128 \\
        activation units & ReLU \\
        optimizer & RMSProp \\
        optimizer learning rate & $2.5\times 10^{-4}$\\
        (RMSProp) squared momentum & 0.95 \\
        (RMSProp) minimum momentum & 0.01 \\
        \hline
    \end{tabular}
\end{table}

\begin{table}[hbt!]
    \centering
    \caption{Parameters used for distributional TAL, MT-DQN and Tsallis-DQN on Atari games. }
    \label{tb:atari}
    \begin{tabular}{llr}
        \cline{1-2}
        Parameter    & Value \\
        \hline
        $T$ (total steps)     &  $3 \times 10^7$     \\
        $C$ (interaction period) &      4     \\
        $|B|$ (buffer size)      & $1 \times 10^6$     \\
        $B_{t}$ (batch size)  &  32 \\
        $\gamma$ (discount rate)       & 0.99      \\
        $I$  (update period)      & $8000$  \\
        $\epsilon$ (epsilon greedy threshold) & $1.0 \rightarrow 0.01 |_{10\%}$ \\
        $\Delta$ (Munchausen number) & $10^{-8}$\\
        $\alpha$ (Tsallis entropy coefficient) &  10 \\
        $\beta$ (advantage coefficient) & 0.9 \\
        Q-network architecture & \\ 
        \multicolumn{2}{c}{\quad $\text{Conv}^{4}_{8,8}32$ - $\text{Conv}^{2}_{4,4}64$ - $\text{Conv}^{1}_{3,3}64$ - FC512 - FC}  \\
        activation units & ReLU \\
        optimizer & Adam \\
        optimizer learning rate & $10^{-4}$\\
        \hline
    \end{tabular}
\end{table}

\subsection{Atari Games}\label{apdx:atari}

We implemented TAL and MT-DQN on top of QR-DQN, which is a famous distributional RL algorithm.
Instead of learning the conditional expectation of cumulative return $Q$ as in \eq{\ref{eq:qvalue}}, distributional RL attempts to learn the distribution itself:
\begin{align}
    \begin{split}
    &Q_{\pi}(s,a) = \mathbb{E}\AdaRectBracket{Z_{\pi}(s,a)},\\
    &\Bellman{}Z(s,a) = r(s,a) + \gamma Z(s',a').
\end{split}
\end{align}
Specifically, QR-DQN aims to approximate $Z$ via quantile regression by parametrizing fixed probabilities (quantiles).
Based on QR-DQN, TAL and MT-DQN are implemented by replacing $Q$ with $Z$ in \eq{\ref{eq:sparse_policy}}.

For the full-fledged Atari games learning is more challenging than the optimized MinAtar games.
We leverage the optimized Stable-Baselines3 architecture \citep{stable-baselines3} for best performance. 
The details can be seen from Table \ref{tb:atari}.
The Q-network uses 3 convolutional layers.
The epsilon greedy threshold is initialized at 1.0 and gradually decays to 0.01 at the end of first 10\% of learning.
For conservative learning, we choose the Tsallis entropy coefficient as $\alpha=10$.

\end{document}